\documentclass[conference]{IEEEtran}
\IEEEoverridecommandlockouts

\usepackage{amsmath,amssymb,amsfonts}
\usepackage{algorithmic}
\usepackage{graphicx}
\usepackage{textcomp}
\usepackage{xcolor}
\usepackage{array}
\usepackage{booktabs}
\usepackage{pifont}
\usepackage{bbding}
\usepackage[backend=biber, sorting=none, style=numeric-comp, maxnames=5]{biblatex}
\usepackage[switch]{lineno}
\usepackage{acronym}
\usepackage{hyperref}

\hypersetup{colorlinks=true, breaklinks=true, pagebackref=true,
  urlcolor=blue, linkcolor=blue,anchorcolor=blue,citecolor=blue,
  hypertexnames=true, final=true,
  pdfpagemode = UseNone, 
  pdfauthor = {},
  pdftitle = {},
  pdfsubject = {},
  pdfkeywords = {}
}

\def\BibTeX{{\rm B\kern-.05em{\sc i\kern-.025em b}\kern-.08em
    T\kern-.1667em\lower.7ex\hbox{E}\kern-.125emX}}

\DeclareSourcemap{
  \maps[datatype=bibtex]{
    \map[overwrite=true]{
      \step[fieldset=url, null]
      \step[fieldset=eprint, null]
      \step[fieldset=doi, null]
    }
  }
}

\acrodef{ADC}[ADC]{Analog-to-Digital Converter}
\acrodef{ADEXP}[AdExp-IF]{Adaptive Exponential Integrate-and-Fire}
\acrodef{ADM}[ADM]{Asynchronous Delta Modulator}
\acrodef{AER}[AER]{Address-Event Representation}
\acrodef{AEX}[AEX]{AER EXtension board}
\acrodef{AE}[AE]{Address-Event}
\acrodef{AFE}[AFE]{Analog Front-End}
\acrodef{AFM}[AFM]{Atomic Force Microscope}
\acrodef{AGC}[AGC]{Automatic Gain Control}
\acrodef{AI}[AI]{Artificial Intelligence}
\acrodef{AMDA}[AMDA]{AER Motherboard with D/A converters}
\acrodef{AMPA}[AMPA]{$\alpha$-Amino-3-hydroxy-5-methyl-4-isoxazolepropionic Acid}
\acrodef{ANN}[ANN]{Artificial Neural Network}
\acrodef{API}[API]{Application Programming Interface}
\acrodef{APMOM}[APMOM]{Alternate Polarity Metal On Metal}
\acrodef{ARM}[ARM]{Advanced RISC Machine}
\acrodef{ASIC}[ASIC]{Application Specific Integrated Circuit}
\acrodef{BCM}[BMC]{Bienenstock-Cooper-Munro}
\acrodef{BD}[BD]{Bundled Data}
\acrodef{BEOL}[BEOL]{Back-end of Line}
\acrodef{BG}[BG]{Bias Generator}
\acrodef{BMI}[BMI]{Brain-Machince Interface}
\acrodef{BTB}[BTB]{Band-to-Band tunnelling}
\acrodef{BTSP}[BTSP]{Behavioral Time Scale Synaptic Plasticity}
\acrodef{CAD}[CAD]{Computer Aided Design}
\acrodef{CAM}[CAM]{Content Addressable Memory}
\acrodef{CAVIAR}[CAVIAR]{Convolution AER Vision Architecture for Real-Time}
\acrodef{CA}[CA]{Cortical Automaton}
\acrodef{CCN}[CCN]{Cooperative and Competitive Network}
\acrodef{CDR}[CDR]{Clock-Data Recovery}
\acrodef{CFC}[CFC]{Current to Frequency Converter}
\acrodef{CHP}[CHP]{Communicating Hardware Processes}
\acrodef{CMIM}[CMIM]{Metal-Insulator-Metal Capacitor}
\acrodef{CML}[CML]{Current Mode Logic}
\acrodef{CMOL}[CMOL]{Hybrid CMOS nanoelectronic circuits}
\acrodef{CMOS}[CMOS]{Complementary Metal-Oxide-Semiconductor}
\acrodef{CNN}[CNN]{Convolutional Neural Network}
\acrodef{CNS}[CNS]{central Nervous System}
\acrodef{COTS}[COTS]{Commercial Off-The-Shelf}
\acrodef{CPG}[CPG]{Central Pattern Generator}
\acrodef{CPLD}[CPLD]{Complex Programmable Logic Device}
\acrodef{CPU}[CPU]{Central Processing Unit}
\acrodef{CSM}[CSM]{Cortical State Machine}
\acrodef{CSP}[CSP]{Constraint Satisfaction Problem}
\acrodef{CTXCTL}[CTXCTL]{CortexControl}
\acrodef{CV}[CV]{Coefficient of Variation}
\acrodef{DAC}[DAC]{Digital to Analog Converter}
\acrodef{DAS}[DAS]{Dynamic Auditory Sensor}
\acrodef{DAVIS}[DAVIS]{Dynamic and Active Pixel Vision Sensor}
\acrodef{DBN}[DBN]{Deep Belief Network}
\acrodef{DBS}[DBS]{Deep Brain Stimulation}
\acrodef{DFA}[DFA]{Deterministic Finite Automaton}
\acrodef{DIBL}[DIBL]{Drain-Induced Barrier-Lowering}
\acrodef{DI}[DI]{Delay Insensitive}
\acrodef{DMA}[DMA]{Direct Memory Access}
\acrodef{DNF}[DNF]{Dynamic Neural Field}
\acrodef{DNN}[DNN]{Deep Neural Network}
\acrodef{DOF}[DOF]{Degrees of Freedom}
\acrodef{DPE}[DPE]{Dynamic Parameter Estimation}
\acrodef{DPI}[DPI]{Differential Pair Integrator}
\acrodef{DRAM}[DRAM]{Dynamic Random Access Memory}
\acrodef{DRRZ}[DR-RZ]{Dual-Rail Return-to-Zero}
\acrodef{DR}[DR]{Dual Rail}
\acrodef{DSP}[DSP]{Digital Signal Processor}
\acrodef{DVS}[DVS]{Dynamic Vision Sensor}
\acrodef{DYNAP}[DYNAP]{Dynamic Neuromorphic Asynchronous Processor}
\acrodef{EBL}[EBL]{Electron Beam Lithography}
\acrodef{ECG}[ECG]{Electrocardiography}
\acrodef{ECoG}[ECoG]{Electrocorticography}
\acrodef{EDVAC}[EDVAC]{Electronic Discrete Variable Automatic Computer}
\acrodef{EEG}[EEG]{Electroencephalography}
\acrodef{EI}[EI]{Excitatory-Inhibitory}
\acrodef{EIN}[EIN]{Excitatory-Inhibitory Network}
\acrodef{EMG}[EMG]{Electromyography}
\acrodef{EM}[EM]{Expectation Maximization}
\acrodef{EOG}[EOG]{Electrooculogram}
\acrodef{EPSC}[EPSC]{Excitatory Post-Synaptic Current}
\acrodef{EPSP}[EPSP]{Excitatory Post-Synaptic Potential}
\acrodef{EZ}[EZ]{Epileptogenic Zone}
\acrodef{FDSOI}[FDSOI]{Fully-Depleted Silicon on Insulator}
\acrodef{FET}[FET]{Field-Effect Transistor}
\acrodef{FFT}[FFT]{Fast Fourier Transform}
\acrodef{FI}[F-I]{Frequency--Current}
\acrodef{FMA}[FMA]{Floating Microelectrode Array}
\acrodef{FNN}[FNN]{Feed-forward Neural Network}
\acrodef{FPGA}[FPGA]{Field Programmable Gate Array}
\acrodef{FR}[FR]{Fast Ripple}
\acrodef{FSA}[FSA]{Finite State Automaton}
\acrodef{FSM}[FSM]{Finite State Machine}
\acrodef{GABA}[GABA]{$\gamma$-Aminobutanoic Acid}
\acrodef{GIDL}[GIDL]{Gate-Induced Drain Leakage}
\acrodef{GOPS}[GOPS]{Giga-Operations per Second}
\acrodef{GPIO}[GPIO]{General Purpose I/O}
\acrodef{GPU}[GPU]{Graphical Processing Unit}
\acrodef{GT}[GT]{Ground Truth}
\acrodef{GUI}[GUI]{Graphical User Interface}
\acrodef{HAL}[HAL]{Hardware Abstraction Layer}
\acrodef{HFO}[HFO]{High Frequency Oscillation}
\acrodef{HH}[H\&H]{Hodgkin \& Huxley}
\acrodef{HMM}[HMM]{Hidden Markov Model}
\acrodef{HRS}[HRS]{High-Resistive State}
\acrodef{HR}[HR]{Heart Rate}
\acrodef{HSE}[HSE]{Handshaking Expansion}
\acrodef{HW}[HW]{Hardware}
\acrodef{ICT}[ICT]{Information and Communication Technology}
\acrodef{IC}[IC]{Integrated Circuit}
\acrodef{IF2DWTA}[IF2DWTA]{Integrate \& Fire 2-Dimensional WTA}
\acrodef{IFSLWTA}[IFSLWTA]{Integrate \& Fire Stop Learning WTA}
\acrodef{IF}[I\&F]{Integrate-and-Fire}
\acrodef{IMU}[IMU]{Inertial Measurement Unit}
\acrodef{INCF}[INCF]{International Neuroinformatics Coordinating Facility}
\acrodef{INI}[INI]{Institute of Neuroinformatics}
\acrodef{IO}[I/O]{Input/Output}
\acrodef{IPSC}[IPSC]{Inhibitory Post-Synaptic Current}
\acrodef{IPSP}[IPSP]{Inhibitory Post-Synaptic Potential}
\acrodef{IP}[IP]{Intellectual Property}
\acrodef{ISI}[ISI]{Inter-Spike Interval}
\acrodef{IoT}[IoT]{Internet of Things}
\acrodef{JFLAP}[JFLAP]{Java - Formal Languages and Automata Package}
\acrodef{LEDR}[LEDR]{Level-Encoded Dual-Rail}
\acrodef{LFP}[LFP]{Local Field Potential}
\acrodef{LIFE}[LIFE]{Longitudinal Intrafascicular Electrodes}
\acrodef{LIF}[LI\&F]{Leaky Integrate-and-Fire}
\acrodef{LLC}[LLC]{Low Leakage Cell}
\acrodef{LNA}[LNA]{Low-Noise Amplifier}
\acrodef{LPF}[LPF]{Low Pass Filter}
\acrodef{LRS}[LRS]{Low-Resistive State}
\acrodef{LSM}[LSM]{Liquid State Machine}
\acrodef{LTD}[LTD]{Long Term Depression}
\acrodef{LTI}[LTI]{Linear Time-Invariant}
\acrodef{LTP}[LTP]{Long Term Potentiation}
\acrodef{LTU}[LTU]{Linear Threshold Unit}
\acrodef{LUT}[LUT]{Look-Up Table}
\acrodef{LVDS}[LVDS]{Low Voltage Differential Signaling}
\acrodef{MCMC}[MCMC]{Markov-Chain Monte Carlo}
\acrodef{MEA}[MEA]{Multielectrode Arrays}
\acrodef{MEMS}[MEMS]{Micro Electro Mechanical System}
\acrodef{MFR}[MFR]{Mean Firing Rate}
\acrodef{MIM}[MIM]{Metal Insulator Metal}
\acrodef{MLP}[MLP]{Multilayer Perceptron}
\acrodef{ML}[ML]{Machine Learning}
\acrodef{MOSCAP}[MOSCAP]{Metal Oxide Semiconductor Capacitor}
\acrodef{MOSFET}[MOSFET]{Metal Oxide Semiconductor Field-Effect Transistor}
\acrodef{MOS}[MOS]{Metal Oxide Semiconductor}
\acrodef{MRI}[MRI]{Magnetic Resonance Imaging}
\acrodef{NCS}[NCS]{Neuromorphic Cognitive Systems}
\acrodef{NDFSM}[NDFSM]{Non-deterministic Finite State Machine}
\acrodef{ND}[ND]{Noise-Driven}
\acrodef{NEF}[NEF]{Neural Engineering Framework}
\acrodef{NHML}[NHML]{Neuromorphic Hardware Mark-up Language}
\acrodef{NIL}[NIL]{Nano-Imprint Lithography}
\acrodef{NI}[NI]{Neural Interface}
\acrodef{NMDA}[NMDA]{\textit{N}-Methyl-\textsc{d}-aspartate}
\acrodef{NME}[NE]{Neuromorphic Engineering}
\acrodef{NN}[NN]{Neural Network}
\acrodef{NOC}[NoC]{Network-on-Chip}
\acrodef{NRZ}[NRZ]{Non-Return-to-Zero}
\acrodef{NSM}[NSM]{Neural State Machine}
\acrodef{OR}[OR]{Operating Room}
\acrodef{OTA}[OTA]{Operational Transconductance Amplifier}
\acrodef{PCB}[PCB]{Printed Circuit Board}
\acrodef{PCHB}[PCHB]{Pre-Charge Half-Buffer}
\acrodef{PCM}[PCM]{Phase Change Memory}
\acrodef{PC}[PC]{Personal Computer}
\acrodef{PDK}[PDK]{Process Design Kit}
\acrodef{PE}[PE]{Phase Encoding}
\acrodef{PFA}[PFA]{Probabilistic Finite Automaton}
\acrodef{PFC}[PFC]{Prefrontal Cortex}
\acrodef{PFM}[PFM]{Pulse Frequency Modulation}
\acrodef{PNI}[PNI]{Peripheral Nerve Interface}
\acrodef{PNS}[PNS]{Peripheral Nervous System}
\acrodef{PPG}[PPG]{Photoplethysmography}
\acrodef{PR}[PR]{Production Rule}
\acrodef{PSC}[PSC]{Post-Synaptic Current}
\acrodef{PSP}[PSP]{Post-Synaptic Potential}
\acrodef{PSTH}[PSTH]{Peri-Stimulus Time Histogram}
\acrodef{PV}[PV]{Parvalbumin}
\acrodef{PYR}[PYR]{Pyramidal}
\acrodef{QDI}[QDI]{Quasi Delay Insensitive}
\acrodef{RAM}[RAM]{Random Access Memory}
\acrodef{RA}[RA]{Resected Area}
\acrodef{RDF}[RDF]{Random Dopant Fluctuation}
\acrodef{RELU}[ReLu]{Rectified Linear Unit}
\acrodef{RISC}[RISC]{Reduced Instruction Set Computer}
\acrodef{RLS}[RLS]{Recursive Least-Squares}
\acrodef{RMSE}[RMSE]{Root Mean Square-Error}
\acrodef{RMS}[RMS]{Root Mean Square}
\acrodef{RNN}[RNN]{Recurrent Neural Network}
\acrodef{ROLLS}[ROLLS]{Reconfigurable On-Line Learning Spiking}
\acrodef{RRAM}[R-RAM]{Resistive Random Access Memory}
\acrodef{RSA}[RSA]{Respiratory Sinus Arrhythmia}
\acrodef{R}[R]{Ripple}
\acrodef{SAC}[SAC]{Selective Attention Chip}
\acrodef{SAT}[SAT]{Boolean Satisfiability Problem}
\acrodef{SCI}[SCI]{Spinal Cord Injury}
\acrodef{SCX}[SCX]{Silicon CorteX}
\acrodef{SD}[SD]{Signal-Driven}
\acrodef{SEM}[SEM]{Spike-based Expectation Maximization}
\acrodef{SLAM}[SLAM]{Simultaneous Localization and Mapping}
\acrodef{SMP}[SMP]{Soil Matric Potential}
\acrodef{SNN}[SNN]{Spiking Neural Network}
\acrodef{SNR}[SNR]{Signal to Noise Ratio}
\acrodef{SOC}[SoC]{System-On-Chip}
\acrodef{SOI}[SOI]{Silicon on Insulator}
\acrodef{SOZ}[SOZ]{Seizure Onset Zone}
\acrodef{SPI}[SPI]{Serial Peripheral Interface}
\acrodef{SP}[SP]{Separation Property}
\acrodef{SRAM}[SRAM]{Static Random Access Memory}
\acrodef{SST}[SST]{Somatostatin}
\acrodef{STDP}[STDP]{Spike-Timing Dependent Plasticity}
\acrodef{STD}[STD]{Short-Term Depression}
\acrodef{STP}[STP]{Short-Term Plasticity}
\acrodef{STT-MRAM}[STT-MRAM]{Spin-Transfer Torque Magnetic Random Access Memory}
\acrodef{STT}[STT]{Spin-Transfer Torque}
\acrodef{SVM}[SVM]{Support Vector Machine}
\acrodef{SW}[SW]{Software}
\acrodef{TCAM}[TCAM]{Ternary Content-Addressable Memory}
\acrodef{TFT}[TFT]{Thin Film Transistor}
\acrodef{TIME}[TIME]{Transverse Intrafascicular Multichannel Electrode}
\acrodef{TLE}[TLE]{Temporal Lobe Epilepsy}
\acrodef{UEA}[UEA]{Utah Electrode Array}
\acrodef{USB}[USB]{Universal Serial Bus}
\acrodef{USEA}[USEA]{Utah Slanted Electrode Array}
\acrodef{VHDL}[VHDL]{VHSIC Hardware Description Language}
\acrodef{VHSIC}[VHSIC]{Very High Speed Integrated Circuits}
\acrodef{VIP}[VIP]{Vasoactive Intestinal Peptide}
\acrodef{VLSI}[VLSI]{Very Large Scale Integration}
\acrodef{VNS}[VNS]{Vagal Nerve Stimulation}
\acrodef{VOR}[VOR]{Vestibulo-Ocular Reflex}
\acrodef{VSA}[VSA]{Vector Symbolic Architecture}
\acrodef{WCST}[WCST]{Wisconsin Card Sorting Test}
\acrodef{WTA}[WTA]{Winner-Take-All}
\acrodef{XML}[XML]{eXtensible Mark-up Language}
\acrodef{divmod3}[DIVMOD3]{Divisibility of a number by three}
\acrodef{hWTA}[hWTA]{Hard Winner-Take-All}
\acrodef{iEEG}[iEEG]{Intracranial Electroencephalography}
\acrodef{rSNN}[rSNN]{recurrent Spiking Neural Network}
\acrodef{sWTA}[sWTA]{soft Winner-Take-All}

\title{A neuromorphic continuous soil monitoring system for precision irrigation}

\author{
    \IEEEauthorblockN{Mirco Tincani\textsuperscript{1,2,*}, Khaled Kerouch\textsuperscript{*,2}, Umberto Garlando\textsuperscript{3},\\
    Mattia Barezzi\textsuperscript{3}, Alessandro Sanginario\textsuperscript{3}, Giacomo Indiveri\textsuperscript{2},\\
    Chiara De Luca\textsuperscript{1,2,\dag}}
    \IEEEauthorblockA{
        \textsuperscript{1}Digital Society Initiative, University of Zurich, Switzerland\\
        \textsuperscript{2}Institute of Neuroinformatics, University of Zurich and ETH Zurich, Switzerland\\
        \textsuperscript{3}Department of Electronics and Telecommunications, Politecnico di Torino, Italy\\
        \textsuperscript{*}These authors contributed equally \quad \textsuperscript{\dag}Corresponding author: chiaradeluca@ini.uzh.ch
    }
}

\begin{document}
%

\maketitle

\IEEEpubidadjcol

\begin{abstract}
  Sensory processing at the edge requires ultra-low power stand-alone computing technologies.
  This is particularly true for modern agriculture and precision irrigation systems which aim to optimize water usage by monitoring key environmental observables continuously using distributed efficient embedded processing elements.
  Neuromorphic processing systems are emerging as a promising technology for extreme edge-computing applications that need to run on resource-constrained hardware.
  As such, they are a very good candidate for implementing efficient water management systems based on data measured from soil and plants, across large fields.
  In this work, we present a fully energy-efficient neuromorphic irrigation control system that operates autonomously without any need for data transmission or remote processing.
  Leveraging the properties of a biologically realistic spiking neural network, our system performs computation, and decision-making locally.
  We validate this approach using real-world soil moisture data from apple and kiwi orchards applied to a mixed-signal neuromorphic processor, and show that the generated irrigation commands closely match those derived from conventional methods across different soil depths.
  Our results show that local neuromorphic inference can maintain decision accuracy, paving the way for autonomous, sustainable irrigation solutions at scale.
\end{abstract}


\section{Introduction}\label{sec:introduction}
Efficient water management is increasingly critical in agriculture, as climate change, population growth, and resource depletion intensify pressure on food systems~\cite{kay2022state}. Precision irrigation addresses these challenges relying on timely, localized measurements of plant and soil conditions, minimizing both under- and over-watering, which can harm yield and crop health.
Recent studies highlight the effectiveness of real-time, on-site monitoring systems for responsive irrigation control~\cite{naeem2023literature, tyagi2024optimizing, obaideen2022overview}. These always-on systems detect early signs of plant stress and trigger prompt corrective actions. Common observables include \ac{SMP} \cite{ben2021automatic}, soil moisture \cite{rios2024iot}, and environmental parameters like air temperature and humidity~\cite{lova2022self}. Different crops and contexts require different sensing strategies and thresholds~\cite{pascoal2024technical}.
A notable example is the WAPPFRUIT project~\cite{barezzi2024wappfruit}, which optimized orchard irrigation using low-power TEROS21 soil matric potential sensors (METER Group, Inc., Pullman, WA, USA) and cloud-connected threshold-based decision logic. This system reduced water usage while maintaining plant health, leveraging LoRaWAN~\cite{mekki2018lorawan} for data transmission and simple rule-based valve control.
However, similar to the WAPPFRUIT project, most existing solutions rely on centralized architectures~\cite{garcia2020iot}, where local sensors transmit data to remote servers via wireless protocols~\cite{gamal2023smart}.
While effective, these systems face challenges such as unreliable connectivity, high energy consumption, and limited scalability—especially in remote or solar-powered deployments~\cite{angelopoulos2020keeping}.
To address these limitations, neuromorphic computing offers a promising alternative~\cite{Covi_etal21,Chicca_etal14}. Inspired by the brain’s efficiency, neuromorphic systems process data in an asynchronous, event-driven fashion, integrating memory and computation in analog or mixed-signal circuits~\cite{Indiveri_Liu15}. This enables ultra-low-power, real-time processing ideal for edge applications like smart irrigation.
In this work, we present a neuromorphic pipeline for real-time irrigation control based on soil moisture sensing. Using a mixed-signal neuromorphic processor and real-world WAPPFRUIT data, we apply the same threshold rules to assess accuracy and responsiveness. By interfacing sensors directly with a biologically inspired, energy-efficient control mechanism, our prototype enables a new class of autonomous, low-power irrigation systems that save water while supporting crop resilience in variable environments.

\section{Methods}\label{sec:methods}

\begin{figure}
    \centering
    \includegraphics[width=1\linewidth]{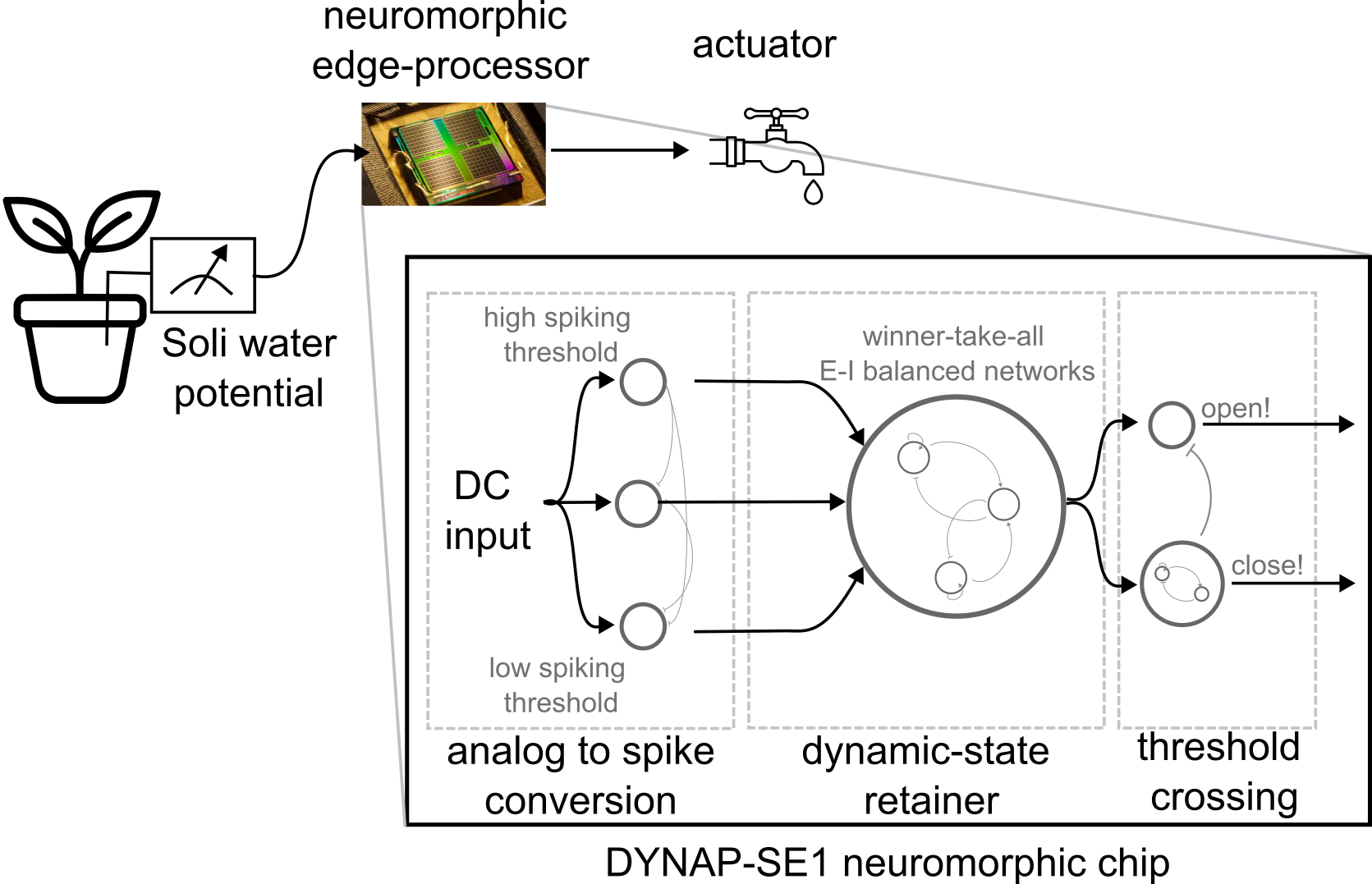}
    \caption{\textbf{Neuromorphic edge-processing pipeline for plant monitoring and water supply management.}
A soil-water potential sensor samples the soil’s moisture state every 15 minutes and sends an analog signal to the DYNAP-SE1 neuromorphic chip.
The signal undergoes analog-to-spike conversion, followed by temporal integration within a dynamic-state retainer—an excitatory-inhibitory balanced, winner-take-all spiking network that preserves slow-varying input dynamics.
A readout mechanism is implemented by connecting populations of spiking neurons such that a “close” signal is triggered by downstream threshold crossings, and an “open” signal by upstream threshold crossings.
These output signals are sent outside the neuromorphic system to actuate a valve, regulating water delivery accordingly.}
    \label{fig:overview}
\end{figure}

The proposed neuromorphic pipeline integrates analog sensing and event-driven neural computation within a compact, low-power architecture (Fig.~\ref{fig:overview}).
Leveraging the properties of mixed-signal neuromorphic circuits present on the DYNAP-SE1 chip~\cite{Moradi_etal18}, we process slow, analog soil moisture dynamic signals, converted into spike-based representations, by a series of brain-inspired spiking neural networks.
The circuits and networks used include an analog-to-spike encoder, a dynamic recurrent network for temporal integration, a winner-take-all neural state machine for stable state retention, and a direction-sensitive readout module that triggers actuator commands.
Each stage is carefully designed to operate under the constraints of variable and noisy ultra-low power circuits implementing co-localized memory and computation by mimicking cortical processing principles, and to produce robust and reliable event-based control.

\subsection{The dataset}\label{sec:method_dataset}
To benchmark our neuromorphic irrigation control system, we used the real-world data presented in the original WAPPFRUIT study~\cite{barezzi2024wappfruit}.
This data consists of a set of time-series measurements of \ac{SMP} collected from drip-irrigated orchards using TEROS21 SMP sensors.
It includes 15-minute interval recordings of \ac{SMP} values, capturing both natural fluctuations due to environmental conditions and irrigation events.
For consistency and reproducibility, we adopted the same irrigation decision thresholds defined in the original paper, where irrigation is triggered, for example in apple orchards, when SMP drops below $-60$\,kPa and is stopped when it rises above $-50$\,kPa, while for kiwi orchards, when SMP drops below $-12$\,kPa and is stopped when it rises above $-5$\,kPa.
These thresholds reflect the optimal soil moisture range for maintaining plant health while minimizing water usage.
This dataset was inverted and normalized to produce values suitable for application to the chip as input currents to the DYNAP-SE1 silicon neurons.
It served as both the input to our neuromorphic network and the ground truth for evaluating the performance of our system in replicating biologically-inspired irrigation decisions.

\subsection{The DYNAP-SE1 chip}\label{sec:method_dynapse}
The neuromorphic processor used in this study is the DYNAP-SE1, a multi-core asynchronous mixed-signal neuromorphic system designed to emulate spiking neuron dynamics in real time~\cite{Moradi_etal18}.
Each of its four cores contains 256 current-mode silicon neurons~\cite{Livi_Indiveri09,Chicca_etal14} that implement faithful models of \ac{ADEXP} neurons~\cite{Brette_Gerstner05}.
Each neuron can receive up to 64 synaptic inputs, configured by programming 64 corresponding \ac{CAM} blocks with the address of the pre-synaptic input source neuron. Each synapse can be further configured to act as excitatory (positive weight) or inhibitory (negative weight), with a fast or slow time constant.
All silicon neurons within the same core share common parameters, including time constants, leak and refractory period values, and support up to 1024 fan-out connections.
Communication among neurons is handled via an \ac{AER} protocol~\cite{Deiss_etal98}, which enables asynchronous event-driven signaling with microsecond-level precision, even under high-load conditions.
In the limit of high input current and with spike-frequency adaptation disabled, the dynamics of the \ac{ADEXP} neuron circuit can be expressed as:
\begin{equation*}
\tau \frac{d}{dt} I_{\text{mem}} + I_{\text{mem}} \approx \frac{I_{\text{in}} I_{\text{gain}}}{I_{\tau}} + \frac{I_a I_{\text{mem}}}{I_{\tau}}
\end{equation*}
where $I_{\text{mem}}$ is the current that represents the membrane potential variable, $\tau$ is the neuron time constant, and the second term on the right-hand side models the positive feedback loop characteristic of the \ac{ADEXP} model.

While direct measurement of DYNAP-SE1 power consumption during online operation is not feasible, it can be estimated as the sum of the energy used for spike generation and communication~\cite{Risi_etal20, Moradi_etal18}, as defined in Eq.~\ref{eq:dynap_power}:

\begin{equation*}
P = \sum_{n=1}^{N} r_n \Bigl(
E_{\text{spike}} + E_{\text{enc}} + N_{\text{cores}}(E_{\text{br}} + E_{\text{rt}})  + N_{\text{cam}}E_{\text{pulse}}
\Bigr)
\label{eq:dynap_power}
\end{equation*}

Here, $r_n$ is the firing rate of neuron $n$, $N$ is the total number of neurons, $N_{\text{cores}}$ is the number of cores each neuron's output is sent to, and $N_{\text{cam}}$ is the number of postsynaptic neurons receiving the spikes.
The energy costs per event, estimated via circuit simulations at 1.8\,V~\cite{Moradi_etal18}, are as follows: spike generation ($E_{\text{spike}}$) and encoding with destination tagging ($E_{\text{enc}}$) each consume 883\,pJ; intra-core broadcasting ($E_{\text{br}}$) consumes 6.84\,nJ; inter-core routing ($E_{\text{rt}}$) consumes 360\,pJ; and extending the output pulse ($E_{\text{pulse}}$) adds 324\,pJ per match.

\begin{figure}
    \centering
    \includegraphics[width=1.\linewidth]{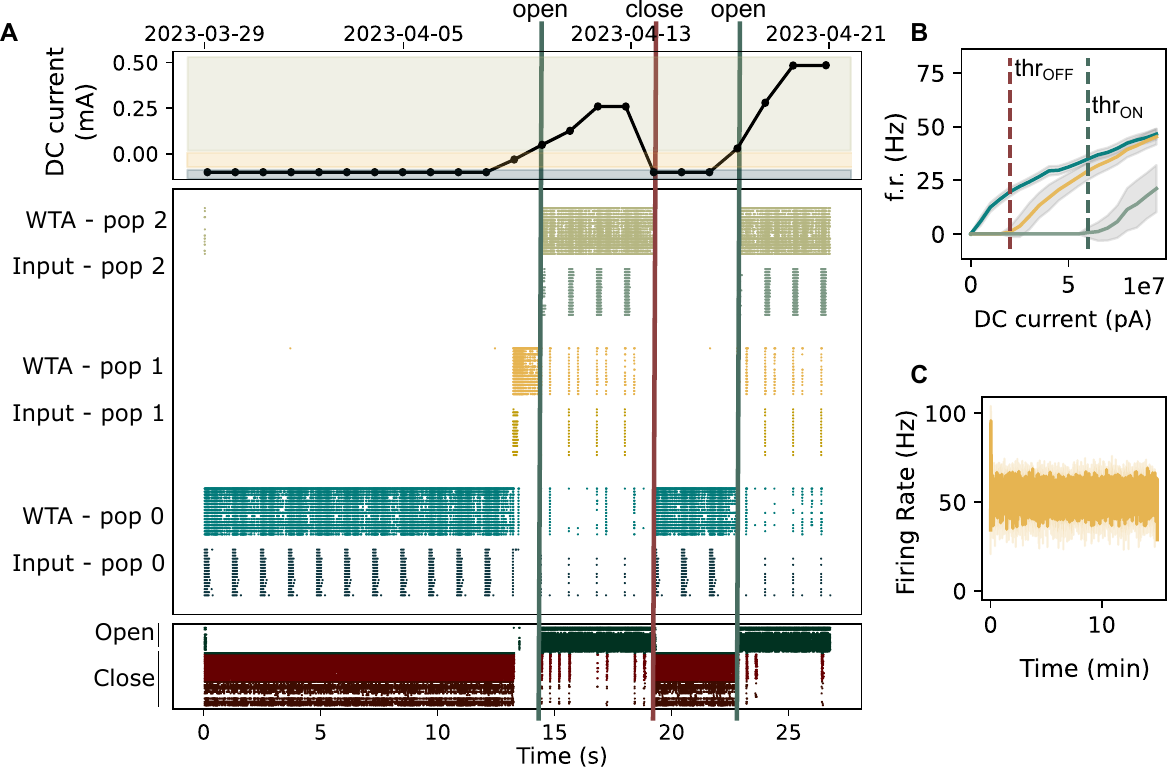}
    \caption{\textbf{Spiking network dynamics for sensor data.} \textbf{A)} \textit{Top row:} Subsampled soil moisture measurements, which serve as inputs to the chip.
Each input, originally recorded every 15\,min, is presented to the system for 200\,ms every 1\,s, leveraging the stable state dynamics of the EI-balanced network.
For visualization purposes, only one data-point per day has been fed to the network.
\textit{Second row:} Raster plots showing spiking activity of the input and \acf{WTA} excitatory neuron populations.
 \textit{Third row:} Activity of the readout layer, indicating the "open" (green) and "close" (red) signals.
\textbf{B)} \acf{FI} curves for three neuron groups, each tuned to get activated at a distinct input DC value.
These activation values are used as opening and closing thresholds in the normalized converted data.
\textbf{C)} Average firing rate of an excitatory \ac{WTA} population across 10 trials in response to a brief (1\,s) stimulus.
The sustained activity over a 15-minute period supports the assumption of a stable state during that time interval.}
    \label{fig:dynamics}
\end{figure}

\subsection{Analog to spike conversion}\label{sec:preprocessing}
To encode \ac{SMP} levels relative to irrigation decision thresholds, we used pulse-frequency modulation implemented on three distinct populations of \ac{ADEXP} neurons, each deployed on a separate core of the DYNAP-SE1 chip. In particular we linearly mapped \ac{SMP} values to synaptic currents, such that \ac{SMP} threshold crossings corresponded to synaptic currents that triggered transitions in our neural system. Indeed, populations were assigned distinct current-injection parameters, tuning their \ac{FI} curves to activate within non-overlapping input ranges. Backward inhibition from each group to those responsive to lower inputs enhanced selectivity, ensuring each population responded to a specific subrange of the rescaled \ac{SMP} signal.
Input signals were linearly rescaled so that transitions between populations aligned with the irrigation activation (\textit{th\textsubscript{ON}}) and deactivation (\textit{th\textsubscript{OFF}}) thresholds used in WAPPFRUIT~\cite{barezzi2024wappfruit}, as shown in Fig. \ref{fig:dynamics}A. Neuron parameters (bias currents and synaptic weights) were adjusted to ensure population-specific activation in response to constant input currents (Fig. \ref{fig:dynamics}B). 
A disinhibitory circuit enforced mutual exclusivity, allowing only the population tuned to the current \ac{SMP} range to spike robustly, despite variability and device mismatch.
The use of entire populations, rather than single neurons, serves two critical purposes: (1) it improves robustness to device mismatch inherent in analog neuromorphic hardware, and (2) it increases the net current delivered to downstream neural state-machine circuits responsible for storing this encoded information in memory.

\subsection{EI balanced dynamics for stable long-term memory}\label{sec:EI}

A core challenge of our system lies in the very slow dynamics in the signals and the 15-minute gap between sensor readings. As our neural processor co-localizes memory and computation,  unlike conventional digital processors, its individual elements can only retain information over timescales set by the circuit dynamics --wich at most can be configured to be in the order of hundreds of milliseconds. To maintain memory over longer timescales, information must be encoded either in the connectivity of the network or in its emergent dynamics.
To address this, we leveraged \ac{EI} balanced dynamics~\cite{Brunel16,Koulakov_etal02}, a well-established mechanism in which excitatory and inhibitory populations tightly interact to maintain dynamic equilibrium within stable attractor states~\cite{Amit92}. This balance prevents runaway excitation and excessive inhibition, allowing the network to remain stable yet responsive, with fluctuation-driven activity and precise temporal behavior.
By tuning the recurrent connectivity and synaptic gains of the \ac{EI} loop, we configured the network to enter a self-sustaining state. After a brief 1\,s, 200\,Hz input, the network maintained a stable firing rate ($53.3 \pm 5$\,Hz) for over 30 minutes without further input. This emergent activity-based memory bridges the temporal gap between sparse sensor updates and continuous computation. All populations preserved their state for more than 15 minutes, as shown in Fig.~\ref{fig:dynamics}C.

\subsection{Neural State Machine}\label{sec:wta}

To encode distinct soil input levels robustly, we deployed a “neural state machine” \cite{Neftci_etal13,Liang_Indiveri19}, a spiking \ac{WTA} network inspired by cortical microcircuits \cite{Rutishauser_Douglas09}. Adapted from~\cite{de2025neuromorphic} for this application, the network has three recurrent excitatory populations ($e_0$, $e_1$, $e_2$), each tuned to different input ranges (Fig.~\ref{fig:dynamics}A).
These populations compete via a shared global inhibitory unit ($\mathit{inh}$) that enforces mutual exclusivity and prevents simultaneous activation. Each excitatory group receives input from a corresponding level-selective neuron (section~\ref{sec:preprocessing}) and forms a persistent attractor state once activated. \ac{EI} balance within populations ensures stable, noise-resistant dynamics.
Activation of one excitatory population suppresses the others through global inhibition, implementing reliable winner-take-all behavior. For faster testing, input intervals were reduced to 1 second during experiments, based on prior results confirming consistent behavior over longer intervals.

\subsection{Direction-based thresholds crossing detector}\label{sec:readout}
Finally, a spiking readout mechanism is implemented using a ``close" state, which maintains persistent activity to signal the actuator accordingly.
In parallel, a distinct ``open" neurons population is activated by strong upward transitions in the input signal and is inhibited by the active ``close" population, ensuring mutual exclusivity and direction-sensitive response.

\section{Results}\label{sec:results}

\subsection{Network response vs threshold detection}\label{sec:res_comparison}
\begin{figure}
    \centering
    \includegraphics[width=1\linewidth]{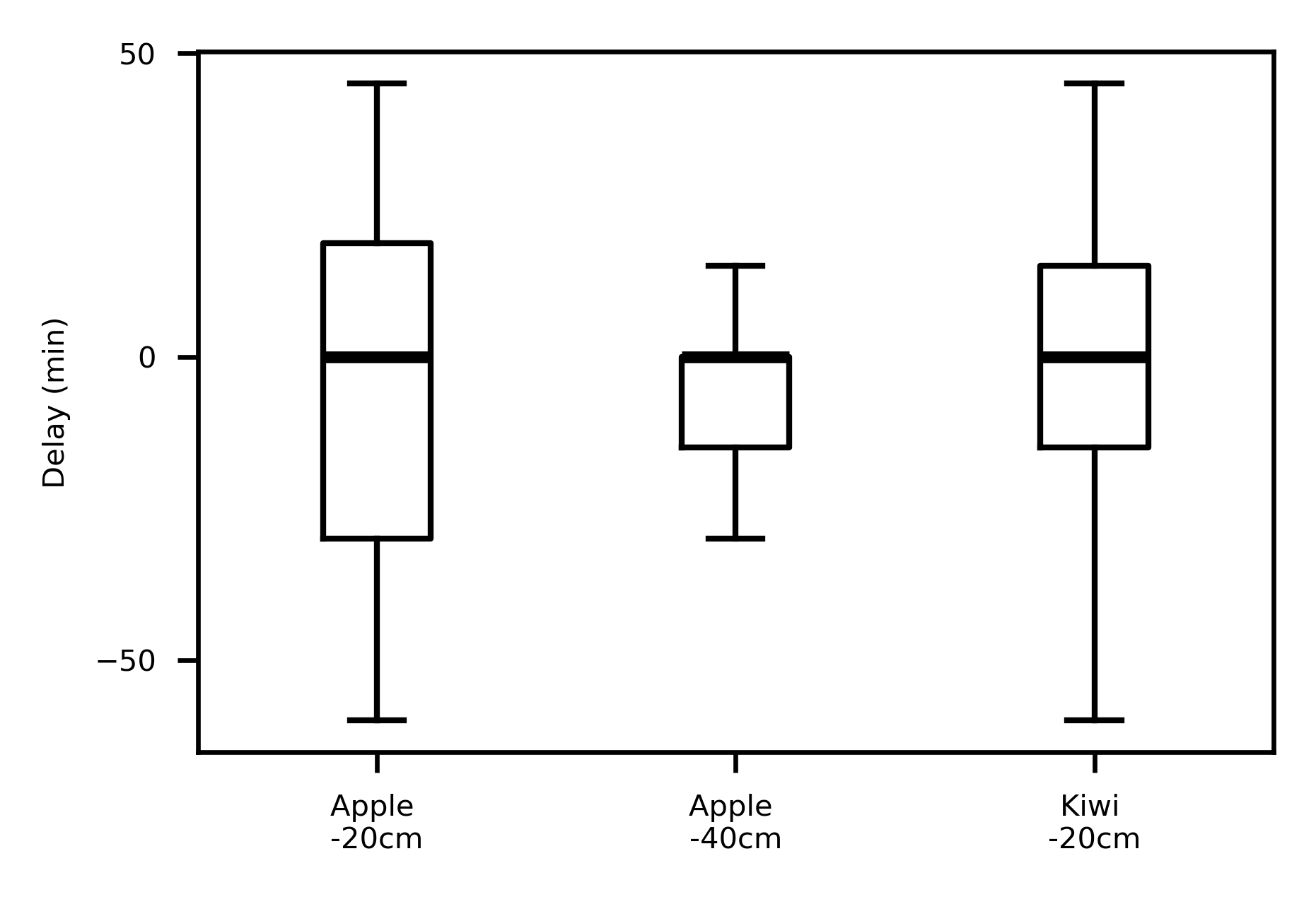}
    \caption{\textbf{Comparison between the open signal of the neuromorphic network and the threshold-based approach.} Time difference between the threshold crossing detected by the neuromorphic system for two different plant species at two different depths. The threshold crossing is defined as the time interval between the sensor signal crossing the activation threshold (\textit{th\textsubscript{ON}}) and the moment when the firing rate of the ``open" neural population exceeds that of the ``closed" population. The line within each box represents the median. Experimental measurements are taken from~\cite{barezzi2024wappfruit}.}
    \label{fig:comparison}
\end{figure}

To evaluate the performance of our system, we compared its irrigation control decisions—specifically the open/close signals—with those that would be generated using the threshold-based approach~\cite{barezzi2024wappfruit} for two different plant types: apples and kiwis.
The comparison was conducted at two soil depths, -20\,cm and -40\,cm.
As shown in Fig.~\ref{fig:comparison}, the discrepancy between the signals is minimal, and the results are highly comparable.
This demonstrates that our edge-based system can reliably replicate the behavior of traditional threshold-crossing methods while avoiding the need for data transmission or remote processing.
These findings support the feasibility of performing irrigation decisions locally, offering both energy and infrastructure advantages.

\subsection{Power consumption}\label{sec:res_power}

\begin{table*}
\centering
\caption{Comparison of IoT irrigation optimization studies}
\begin{tabular}{
    >{\raggedright\arraybackslash}p{2cm}
    | >{\raggedright\arraybackslash}p{2.5cm}
    | >{\raggedright\arraybackslash}p{2.5cm}
    | >{\raggedright\arraybackslash}p{2.5cm}
}
\toprule
\textbf{Parameter} & \textbf{This work} & \textbf{Barezzi et al. (2024)~\cite{barezzi2024wappfruit, barezzi2022}} & \textbf{Angelopoulos et al. (2020)~\cite{angelopoulos2020keeping}} \\
\midrule
\midrule
\textbf{On-the-edge} &
\checkmark &
x &
x \\
\midrule
\textbf{Data used} &
Data from Barezzi et al. 2024~\cite{barezzi2024wappfruit} &
Apple/Kiwi orchards &
Strawberry greenhouses \\
\midrule
\textbf{Energy consumption} &
5.97\,$\mu Wh$  &
11.39\,$\mu Wh$ &
1.44\,$mWh$ \\

\midrule
\textbf{Resting energy} &
5.96\,$\mu Wh$ &
1.89\,$\mu Ah$ @3.6\,$V$ = 6.8\,$\mu Wh$ &
0.05\,$mAh$ @3.3\,$V$ = 0.165\,$mWh$ \\
\midrule
\textbf{Active energy} &
0.003\,$\mu Wh$ &
1.28\,$\mu Ah$ @3.6\,$V$ = 4.59\,$\mu Wh$ &
0.389\,$mAh$ @3.3\,$V$ = 1.28\,$mWh$ \\

\bottomrule
\end{tabular}
\label{tab:power}
\end{table*}

Based on the approach described in section~\ref{sec:method_dynapse} values, we estimated the energy required to process a single input using the DYNAP-SE1 chip over a 200~ms activation window every 15 minutes.
This results in a power consumption of approximately 5.97\,$\mu Wh$ per event.
 Table~\ref{tab:power} presents a comparative analysis of energy consumption across recent IoT-based irrigation optimization studies (without addressing the energy consumption of the sensors).
It is important to note that the energy figures reported for the other works account only for the transmission of data and do not include the energy costs associated with the inference processing, which are typically carried out on remote servers.

In contrast, our approach executes the entire inference pipeline directly on the edge device, eliminating the need for data transmission. Communication is required only when a command must be sent to the irrigation system.
This design results in drastically lower active energy consumption—only 0.003\,$\mu$Wh—while enabling fast and efficient on-device decision-making.
By keeping all processing local, our system not only reduces latency but also achieves significant energy savings, making it highly suitable for large-scale, energy-constrained agricultural deployments.

\section{Conclusions}\label{sec:conclusions}

We presented a fully neuromorphic, ultra-low-power system for autonomous, edge-based irrigation control, eliminating the need for wireless transmission or cloud inference. Leveraging the DYNAP-SE1 chip, our spiking neural pipeline translates sparse analog inputs into stable actuator commands with only 5.97\,$\mu$Wh total energy consumption.
%
The system integrates memory, computation, and decision-making on a single chip. Tested on real-world soil moisture data, it closely replicates threshold-based irrigation strategies across crop types and depths, demonstrating accuracy and robustness.
%
To handle challenges related to long-timescale processing due to sparse sensor update, we used stable EI-balanced dynamics for memory retention and conservative state transitions, thus ensuring reliable decisions in dynamic conditions.
%
Though already effective as it is, this system can scale to multi-modal data by converting diverse sensor inputs into spikes and processing them locally on neuromorphic hardware—enabling classification and control without cloud resources. This feasibility study highlights the promise of fully local, energy-autonomous control for precision agriculture.
Next, we plan to integrate neuromorphic chips with physical sensors and actuators for in-field deployment, validating system performance under real-world variability. We also aim to support multi-sensor integration to enhance system versatility and confirm neuromorphic edge computing as a practical tool for sustainable agriculture.

\section*{Acknowledgment}
C.D.L. and M.T. acknowledge the financial support of the Bridge Fellowship founded by the Digital Society Initiative at University of Zurich (Grant No. G-95017-01-12).

\renewcommand{\bibfont}{\small} 
\printbibliography

\end{document}